\documentclass{article}
\usepackage{PRIMEarxiv}
\usepackage[utf8]{inputenc} 
\usepackage[T1]{fontenc}    
\usepackage{hyperref}       
\usepackage{url}            
\usepackage{booktabs}       
\usepackage{amsfonts}       
\usepackage{nicefrac}       
\usepackage{microtype}      
\usepackage{lipsum}
\usepackage{fancyhdr}       
\usepackage{graphicx}       
\graphicspath{{media/}}     
\usepackage{float}
\title{PPG Signals for Hypertension Diagnosis: A Novel Method using Deep Learning Models
}

\author{
  Graham Frederick \\
  School of Electronics Engineering \\
  Vellore Institute of Technology \\
  Chennai, India\\
  \texttt{graham.frederick2019@vitstudent.ac.in} \\
   \And
  Yaswant T \\
  School of Electronics Engineering \\
  Vellore Institute of Technology \\
  Chennai, India\\
  \texttt{yaswant.t2019@vitstudent.ac.in} \\
  \And
  Brintha Therese A  \\
  School of Electronics Engineering \\
  Vellore Institute of Technology \\
  Chennai, India\\
  \texttt{abrinthatherese@vit.ac.in} \\
}

\begin{document}
\maketitle

\begin{abstract}
Hypertension is a medical condition characterized by high blood pressure, and classifying it into its various stages is crucial to managing the disease. In this project, a novel method is proposed for classifying stages of hypertension using Photoplethysmography (PPG) signals and deep learning models, namely AvgPool\_VGG-16. The PPG signal is a non-invasive method of measuring blood pressure through the use of light sensors that measure the changes in blood volume in the microvasculature of tissues. PPG images from the publicly available blood pressure classification dataset were used to train the model. Multiclass classification for various PPG stages were done. The results show the proposed method achieves high accuracy in classifying hypertension stages, demonstrating the potential of PPG signals and deep learning models in hypertension diagnosis and management.
\end{abstract}

\keywords{PPG \and Multiclass classification \and pooling layers}

\section{Introduction}
According to the World Heart Federation, 1.3 billion people worldwide were diagnosed with the condition of hypertension as of 2022, and around 10 million succumb to it every year. It is quite necessary that a person have a heart-healthy lifestyle to prevent the risk of getting diagnosed with fatal diseases like cardiovascular disease(CVD) and arterial diseases. If a person is negligent about monitoring his or her blood pressure on a regular basis,they may get diagnosed with it suddenly. It is difficult to anticipate hypertension at times,as the symptoms in the initial stages of this condition don't feel significant. The classification of stages of hypertension also comes in handy so that a person can make the following lifestyle changes based on the stage he is classified in. 
Some of the conventional methods of blood pressure measurement included cuff based blood pressure monitoring. As of present, there have been several machine learning and deep learning algorithms developed to learn about the risk of hypertension beforehand.
 
Photoplethysmography (PPG) is an optical method of detecting changes in the circulation of blood. Mainly,it is a waveform that denotes variations in blood volume over time. Some of the very common applications of PPG signals are heart rate measurement and blood flow monitoring.
PPG signals have recently become famous due to their non-invasive nature,convenient nature,inexpensive method for estimating blood pressure, and huge potential in wellness monitoring \cite{1}. Blood pressure can be estimated indirectly by calculating the time delay between the starting part of the systolic phase and the upstroke of the arterial pulse wave. Features extractable from PPG signals include a diastolic peak, a systolic peak and a notch in the middle \cite{2}.Research has proved that estimating few features of PPG signals tend to be difficult,as the traits of PPG differ from person to person\cite{3}.
In this study, we propose a multiclass classifier model to classify stages of hypertension. Y Liang et al. collected data on 219 patients, as reported in their study \cite{4}. The data is pre-processed and the waveform image will be used to train the models. The results will set out to improve the performance and accuracy in the classification models of hypertension using PPG signals.

\section{Literature Review}
Majid Nour and Kemal Polat \cite{5} have created various classification algorithms and tested them using a 5 fold cross validation technique. PPG signals were used to extract some information to evaluate diseases with the proposed models.The authors made use of classification algorithms like the C4.5 decision tree classifier, linear discriminant analysis (LDA), linear support vector machine (LSVM) and random forest. These models results in classification were compared to one another. The results were concluded based on the confusion matrix, ROC curve and F-measure value and inferred that the C4.5 decision tree and random forest are the best methods that can be implemented to classify stages of hypertension.

Fabian Schrumpf et al. \cite{6} have studied the various regression and classification deep neural network algorithms used for blood pressure prediction. The proposed models included a modified version of AlexNet and three versions of ResNet: ResNet18,ResNet34 and ResNet50. Personalization has been implemented to fine-tune the results. The authors found significant differences in results before and after personalization and concluded that a classification based approach is a little more precise and can be used over a regression based approach.

Martha Pulido et al. \cite{7} decided to implement a modular neural network on systolic, diastolic blood pressure and classify the hypertension stage of a patient . A modular neural network was created for systolic, diastolic and pulse rate modules and the best architecture was found with the first layer having 26 neurons and the second layer having 29 neurons. A statistical study and Z test were done for all modules, and the error obtained by all 3 modules was less than the error obtained by the regression method, providing good results.

Erick Martinez-Ríos et al. \cite{8} have studied the various machine learning models created to estimate blood pressure using physiological data. The studies regarding classification that use clinical data or try to estimate BP values have standard variables while the models that use PPG and ECG wave data use feature extraction and proceed to train the model. The authors discussed the numerous limitations and issues, including the lack of research on artifact removal in physiological signals and the necessity for learning models that could deal with the overfitting problem.

Xiaoxiao Sun et al. \cite{3}  have explored the use of PPG signals and its two derivatives to predict blood pressure. The Hilbert-Huang Transform (HHT) method was used to process the PPG signal. Four types of deep learning methods, such as AlexNet,GoogleNet,ResNet18 and ResNet34 were used on the HHT transformed dataset. These datasets have been put into the four different CNN layers for training. It was concluded that increasing the number of convolution layers does not necessarily result in a proportional improvement in the outcomes.

Chih-Ta Yen et al. \cite{9} created a deep learning neural network that uses the ResNetCNN and BiLSTM models along with a neural network made of Xception and BiLSTM to classify hypertension into four stages. Models were created with various layers and kernel sizes, and accuracy, recall, and precision were calculated. It is found that the Xception + BiLSTM model with 37 layers, 32 kernels, and kernel size 36 produced the best results when compared to ResNetCNN +BiLSTM. A confusion matrix of the best model results was made to show the classification results.

Yunendah Nur Fuadah and Ki Moo Lim \cite{10} proposed a 1D concatenated CNN model that uses PPG and ECG signals to classify blood pressure into five levels based on systolic and diastolic values according to American Heart Association standards. MIMIC-III database was used to obtain PPG and ECG signals. The signals were sent to the 5 CNN architectures, each with different layers and evaluation is done by calculating its accuracy, precision, recall, F1 score and AUC. The concatenated model with five convolution layers produced the best results and an accuracy of 95.

Ali Bahari Malayeri and Mohammad Bagher Khodabakhshi \cite{11} used a fuzzy recurrence plot function to change the PPG signal into a fuzzy recurrence plot and sent it to a 2D Convolution neural network, concatenating those features with the same PPG signal features obtained from a 1D CNN model. The MIMIC-II dataset was used to obtain the PPG signals, and three different CNN models were proposed: 1D\_CNN, 2D\_CNN, and Concat\_CNN, which consists of the first two models. The Concat\_CNN model performed best, with the least mean absolute error and satisfying the BHS and AAMI standards.

Y H Tanc and M Ozturk \cite{12} in their study performed a synchrosqueezing transform to obtain a time varying spectra of PPG signals. A Googlenet CNN model was implemented to classify the signal into normotension and hypertension and performance metrics such as accuracy, sensitivity, specificity and F1 score were used to evaluate the models. With 80\% training data and the rest 20\% used for testing, the model achieved an accuracy of 95.8\% with 96.8\% F1-score.

Jesus Cano et al. \cite{13} conducted a study in which they performed wavelet transform on PPG signals and made two classifications based on the results : normotension vs. prehypertension + hypertension and normotension + prehypertension vs. hypertension. To evaluate the models, they used pretrained Googlenet and ResNet architectures and considered evaluation metrics such as sensibility, specificity, and F1 score. The study also involved intra-patient classification using different PPG signals. The highest F1 score achieved was 90.28 under the Googlenet architecture.

Filipa Esgalhado et al. \cite{14} aimed to classify and process PPG signals by finding the time limit for each PPG beat to identify the peak of the PPG waveform. To accomplish this, they proposed the use of the BiLSTM and CNN-LSTM models on PPG signals, as well as the Synchrosqueezed Fourier Transform (SSFT) on PPG signals. In addition to accuracy, they considered evaluation metrics such as precision, recall, and F1 scores. The CNN-LSTM model with SSFT PPG signal achieved the highest performance, with a precision of 92.3\% and an accuracy of 89.4\%.

Qian Wu \cite{15} studied the important features extractable from PPG signals and classified PPG waveforms for healthy and unhealthy participants. They proposed an improved K-means algorithm for feature extraction based on systolic area, diastolic area and the entire pulse wave area and used a probability neural network (PNN) for classification. Their evaluation metric was the confidence score, which achieved a score of 95.57\%. They achieved an accuracy of 80\% for classifying unhealthy patients based on 13 training input samples and 10 classification samples.

These studies demonstrate that various techniques can be employed to classify and process PPG signals, and the choice of algorithm and evaluation metric can have a significant impact on the performance of the model.

\section{Methodology}
\subsection{Dataset Description and Pre-processing}
\begin{figure}[H]
    \centering
    \includegraphics[height=50mm,scale=0.25]{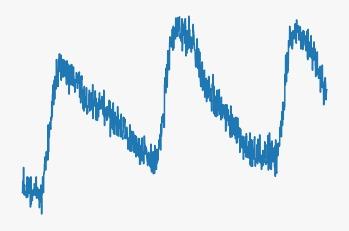}
    \caption{Signal before pre-processing}
    \label{fig:Rawdata}
\end{figure}

\begin{figure}[H]
    \centering
    \includegraphics[height=50mm,scale=0.25]{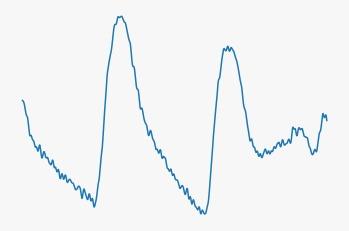}
    \caption{Signal after pre-processing}
    \label{fig:processeddata}
\end{figure}

\begin{figure}[H]
    \centering
    \includegraphics[height=75mm,scale=0.25]{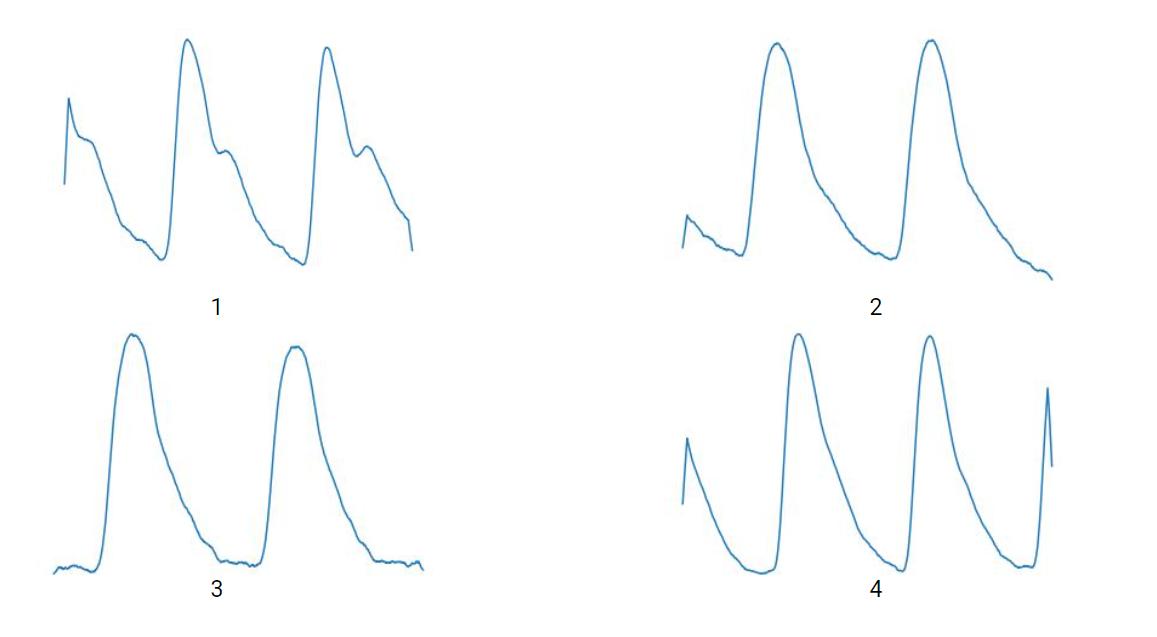}
    \caption{1:normal 2:prehypertension  3:hypertension stage-1 4:hypertension stage-2}
    \label{fig:PPGimage}
\end{figure}

Y Liang et al. \cite{4} created a medical dataset that contains clinical and PPG signal information for 219 patients. The PPG signals of each patient were divided into three time periods, each with a time range of 2.1 seconds and 2100 samples. A total of 657 PPG signals are available with the patient’s clinical information. The PPG signals can be classified into 4 hypertension stages with the help of the clinical information provided as shown in Figure \ref{fig:PPGimage}. There are 246 normal (nt) , 255 pre-hypertension (pt), 99 stage 1 hypertension (ht1), 57 stage 2 hypertension (ht2) PPG signals available in this dateset.
Figure \ref{fig:Rawdata} shows the raw data provided by the dataset. Pre-processing was done to remove noise from the PPG signals. Moving average method was used with window size of 50 to ensure signal data is not lost at the same time smoothing the curve to make it fit for training. The resulting image after pre-processing is displayed in Figure \ref{fig:processeddata}.

\subsection{Learning Models}

Several machine learning methods were proposed for the classification of the four stages of hypertension based on the dataset described above.The classification methodologies adapted here were: Alexnet, Resnet -50, VGG-16 and the novel model: AvgPool\_VGG-16. These algorithms were put to use keeping the datasets volume and variety in mind. The methods used were elaborated in detail in the following sections.

\subsubsection{AlexNet}

Alexnet is a convolutional neural network model introduced by Alex Krizhevesky, Ilya Sutskever and Geoffrey Hinton in 2012 \cite{29}. Alexnet’s structure consists of eight layers. An 11x11 convolution filter is present in the first layer, which is followed by a pooling process. The same max pooling operations are done to the second filter of size (5x5) and the other three filters of size (3x3). The final three filters are hidden layers with dropouts, and a softmax function is used at the end \cite{16}. It is considered a groundbreaking architecture in the field of deep learning, which is a significant improvement over the conventional deep learning models \cite{17}.
Its contributions to deep learning are immense, as it helped to establish the effectiveness of deep neural networks in computer vision tasks and paved the way for the development of even more powerful and complex neural network architectures.

\subsubsection{ResNet-50}

ResNet-50 is a deep neural network architecture that was introduced in 2015 by researchers Kaiming He, Xiangyu Zhang, Shaoqing Ren and Jian Sun \cite{32}. It is a variant of the Residual Network (ResNet) architecture, which is based on the idea of residual learning. Residual learning involves the use of skip connections that allow the input of one layer to be directly passed to a layer beyond the next, rather than going through a series of intermediate layers. This helps to alleviate the vanishing gradient problem, which can occur in very deep networks. Its performance has been shown to be superior to many previous state-of-the-art models\cite{18}.

\subsubsection{VGG-16}

\begin{figure}[h]
    \centering
    \includegraphics[height=100mm,scale=1.00]{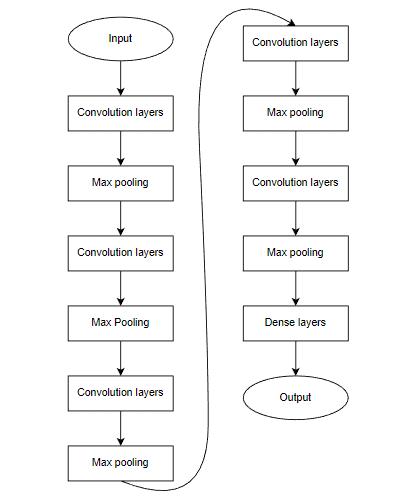}
    \caption{VGG-16 architecture}
    \label{fig:VGG}
\end{figure}

Visual Geometry Group-16 is a type of convolutional neural network architecture, that consists of 13 convolutional layers and 3 fully connected layers \cite{19}. When the input image is passed into this architecture, it gets passed onto a convolutional layer and gets applied to a filter with certain parameters in order to extract features. The output of this convolutional layer gets passed on to the other convolutional layers and is applied with ReLu ,an activation function that is faster to compute. Then the output goes to the fully connected layers and the softmax layer, where the classification begins based on the features extracted. This model was first developed by Karen Simonyan and Andrew Zisserman in 2014. It is mainly implemented in classification tasks like image classification. VGG-16 proved to be a very important accomplishment as it performed better than older models like Alexnet thanks to its many layers with smaller kernels, which produced excellent accuracy. Figure \ref{fig:VGG} shows us the existing VGG-16 architecture. After each convolution layer, the output is passed through a max pooling layer that extracts the image features and is finally sent to the dense layers.

\subsubsection{AvgPool\_VGG-16}

\begin{figure}[h]
    \centering
    \includegraphics[height=100mm,scale=1.00]{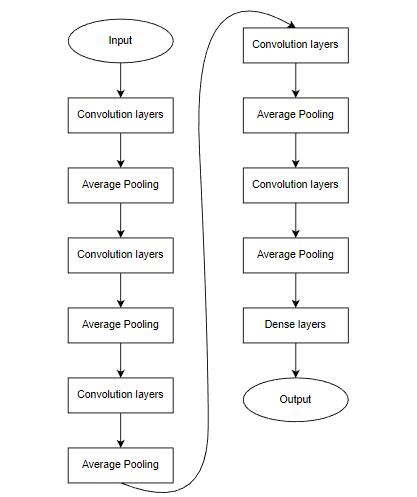}
    \caption{Proposed AvgPool\_VGG-16 architecture}
    \label{fig:avgVGG}
\end{figure}

Average Pooling is a widely used operation in convolutional neural networks that computes the average of patches of a feature map and is used to reduce the size of it. This method sets out to decrease the parameters in the model to increase its robustness and efficiency. Compared to Max pooling, it extracts features easily \cite{20}.
Just like average pooling, max pooling is another common method used to diminish the size of feature maps. It extracts more features from the input image than average pooling. It also obtains the most important features, by taking only the maximum value of each region.
In machine learning, the choice of pooling method can significantly impact the performance of the classifier. Based on the input, the pooling layers have their own drawbacks\cite{21}. For physiological signals, using average pooling is better than max pooling since average pooling has the ability to preserve temporal information unlike max pooling, which saves only the maximum value and discards temporal information that is crucial in classification\cite{22}. Furthermore, average pooling has the ability to reduce noise as it has a noise suppression mechanism\cite{23}. Max pooling takes only edges/peak values into consideration. Since the signals are normalized from 0 to 1, max pooling becomes less efficient than average pooling. Figure \ref{fig:avgVGG} shows the proposed model architecture where we chose to use average pooling instead of max pooling in our VGG-16 model as it yields better results. 

\section{Evaluation Metrics}

When evaluating the effectiveness of a classification model, precision and recall are two metrics that are often used. Precision is calculated as the ratio of the number of images correctly classified as belonging to the positive class to the total number of images classified as belonging to the positive class by the model \cite{24}. The Recall metric measures the percentage of actual positive cases that the model correctly identifies \cite{30}.
The F1 score is a significant evaluation metric in machine learning that presents the weighed harmonic mean of precision and recall scores to determine the model's overall performance on the dataset \cite{25}. It indicates the frequency of accurate predictions made by the model throughout the dataset \cite{26}. 
Support is a metric that indicates the number of times each class appears in true responses. The measure values for each class are added together, and the support for each class is weighted based on its frequency \cite{27}. The weighted average is then calculated accordingly.
Accuracy is an evaluation metric that is used for classification models to determine the proportion of correct predictions made by the model among all the predictions made \cite{30}. However, accuracy is only an appropriate metric when the dataset has a balanced distribution of classes. In other words, when the number of instances of each class in the dataset is roughly equal, accuracy can be a useful metric to assess the performance of the model \cite{28}.The metrics are represented as \ref{eq:1} - \ref{eq:5} where TP: True Positive TN: True Negative FP: False Positive FN : False negative \cite{31}

\begin{equation}\label{eq:1}
Precision= TP / (TP + FP)
\end{equation}

\begin{equation}\label{eq:2}
Recall = TP / (TP + FN)
\end{equation}

\begin{equation}\label{eq:3}
F1 score  = 2 * (precision * recall ) / (precision + recall)
\end{equation}

\begin{equation}\label{eq:4}
Weighted Average = summation (Metrics * support) /summation(Support)    
\end{equation}

\begin{equation}\label{eq:5}
Accuracy= (TP + TN)/(TP+FN+FP+TN)
\end{equation}

\section{Results}

\begin{figure}[H]
    \centering
    \includegraphics[height=100mm,scale=1.00]{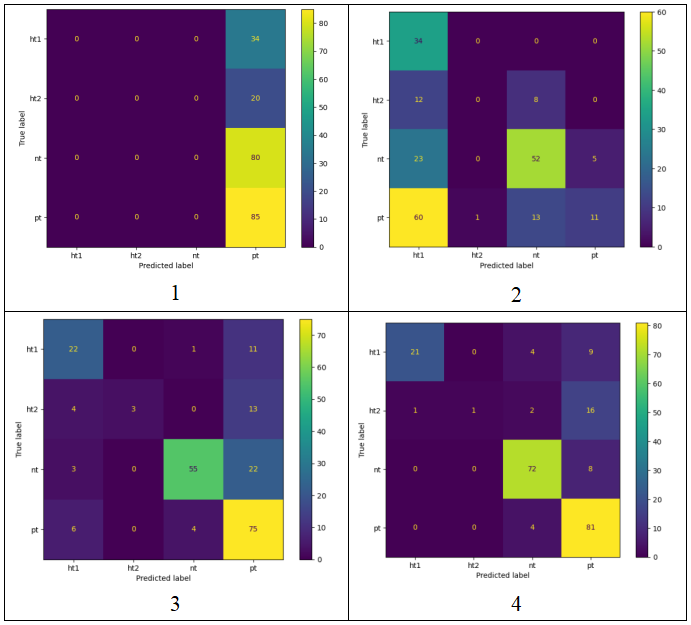}
    \caption{Confusion Matrix for 1)Alexnet 2) Resnet-50 3)VGG-16 4)AvgPool\_VGG-16}
    \label{fig:confmatrix}
\end{figure}

Using data augmentation, a test dataset was created by adding and removing noise to the PPG signals and is used for validation. The deep learning models were all trained with 100 epochs on 20 epochs of data to prevent overfitting.
All models were trained , and accuracy, F1 score , and confusion matrix were determined based on the results obtained and compared with the test results. It is observed that the models aren't able to distinguish between stage 1 hypertension and stage 2 hypertension due to data insufficiency and since the signals have minor changes. Since these PPG signals are recorded and pre-processed from the patients [4] rather than simulated by a system , it is necessary to train the models in such datasets so they will be able to predict in real life scenarios. 
Figure \ref{fig:confmatrix} provides us the confusion matrix of the models and helps us find true positives (TP), true negatives (TN), false positives (FP), and false negatives (FN) value for each class.

Alexnet produced the least accuracy and wasn't able to classify any signal according to Figure \ref{fig:confmatrix}. Pre-trained ResNet and VGG-16 models are implemented. While ResNet-50 provides better results than Alexnet, the VGG-16 model has produced significantly better results than all the other models with an accuracy of 71\% and an F1 score of 0.69. Our modified VGG model, AvgPool\_VGG-16, which utilizes average pooling, achieved the best results with an accuracy of 80\% and an F1 score of 0.77. 

\begin{table}[H]
\centering
\caption{Evaluation of Deep learning models}
\label{tab:eval}
\begin{tabular}{|l|l|l|l|l|l|l}
\hline
\textbf{Model} & \textbf{Class} & \textbf{Precision} & \textbf{Recall} & \textbf{F1 score} & \textbf{Support} \\ \hline
Alexnet& ht1& 0.00& 0.00& 0.00 & 34 \\
       & ht2 & 0.00 & 0.00 & 0.00 & 20 \\
       & nt & 0.00 & 0.00 & 0.00 & 80 \\
       & pt & 0.39 & 1.00 & 0.56 & 85 \\ \cline{2-6} 
       & Accuracy & &  & 0.39 & 219 \\
       & Weighted average  & 0.15 & 0.39 & 0.22 & 219 \\ \hline
Resnet-50& ht1 & 0.26 & 1.00 & 0.42 & 34 \\
        & ht2 & 0.00 & 0.00 & 0.00 & 20 \\
       & nt & 0.71 & 0.65 & 0.68 & 80 \\
       & pt & 0.69 & 0.13 & 0.22 & 85 \\ \cline{2-6} 
       & Accuracy & & & 0.44 & 219 \\
       & Weighted average  & 0.57 & 0.44 & 0.40 & 219 \\ \hline
VGG-16 & ht1 & 0.63 & 0.65 & 0.64 & 34 \\
       & ht2 & 1.00 & 0.15 & 0.26 & 20 \\
       & nt & 0.92 & 0.69 & 0.79 & 80 \\
       & pt & 0.62 & 0.88 & 0.73 & 85 \\ \cline{2-6} 
       & Accuracy & & & 0.71 & 219 \\
       & Weighted average  & 0.76 & 0.71 & 0.69 & 219 \\ \hline
AvgPool\_VGG-16& ht1  & 0.95 & 0.62 & 0.75 & 34 \\
       & ht2 & 1.00 & 0.05 & 0.10 & 20 \\
       & nt & 0.88 & 0.90 & 0.89 & 80 \\
       & pt & 0.71 & 0.95 & 0.81 & 85 \\ \cline{2-6} 
       & Accuracy &  & & 0.80 & 219 \\
       & Weighted average  & 0.84 & 0.80 & 0.77 & 219 \\ \hline     
\end{tabular}
\end{table}

\begin{table}[H]
\centering
\caption{Model performance comparison}
\label{tab:performance}
\begin{tabular}{|l|l|l|l|}
\hline
Model & Accuracy & F1 Score \\
\hline
Alexnet & 39 & 0.22 \\
Resnet 50 & 44 & 0.40 \\
VGG-16 & 71 & 0.69 \\
Xception+BiLSTM [9] & 76 & 0.48 \\
AvgPool\_VGG-16 & 80 & 0.77 \\
\hline
\end{tabular}
\end{table}

\section{Discussion}

Table \ref{tab:eval} presents the results obtained from running the models on a fixed test dataset. AlexNet provided poor results and was unable to distinguish the signals. ResNet-50 was able to classify normal PPG signals, but it was not able to distinguish the other classes. VGG-16 and AvgPool\_VGG-16 provided good results, but their accuracy was lower when classifying stage-2 hypertension (ht2) because there were relatively few data records available. 
From Table \ref{tab:performance}, we are able to find that the VGG-16 model has produced better results than the Alexnet and ResNet. Since VGG-16 has more training parameters than ResNet and AlexNet, it has produced higher accuracy results than the other models when it comes to medical signals. The results obtained from our proposed model, AvgPool\_VGG-16, demonstrate that average pooling is a more suitable pooling technique for signal image classification than max pooling. Additionally, this model has provided better results compared to previously proposed models \cite{9}. Our model achieved the best performance, indicating that the use of average pooling offers advantages over max pooling in this context. This highlights the importance of selecting appropriate pooling techniques for signal image classification tasks.

\section{Conclusion and Future works}

In conclusion, this paper presents a novel approach for classifying PPG signals based on a 2.1s time window using our proposed AvgPool\_VGG-16 model. The results show that the proposed model outperforms traditional models such as VGG-16, AlexNet, and ResNet in PPG signal classification. The results of this study suggest that the proposed approach is a promising method for PPG signal classification. However, due to the smaller number of records in the dataset, the study can only provide insights into how pooling layers impact signal classification and its results as compared to other deep learning models.
The accuracy obtained can be improved by increasing the dataset by adding more subjects. It is possible for the models to produce better results when pre-processed by using different medical signal transforms to spread information along the whole image rather than having it be a wave. 
It can be recommended that future studies explore the use of other pooling techniques to further improve the accuracy of PPG signal classification models. Additionally, it would be useful to investigate the impact of varying the size of the pooling filters and how it affects accuracy.

\end{document}